\documentclass[conference]{IEEEtran}
\IEEEoverridecommandlockouts
\usepackage[noadjust]{cite}

\usepackage{soul}
\usepackage{tcolorbox}
\usepackage{ragged2e}
\usepackage{tabularx}
\usepackage{hyperref}

\usepackage{amsmath,amssymb,amsfonts}
\usepackage{graphicx}
\usepackage{textcomp}
\usepackage{xcolor}
\usepackage{algorithm}
\usepackage{algpseudocode}
\usepackage{caption}

\usepackage{multicol}
\usepackage{multirow}
\usepackage{array}
\usepackage{balance}

\def\BibTeX{{\rm B\kern-.05em{\sc i\kern-.025em b}\kern-.08em
    T\kern-.1667em\lower.7ex\hbox{E}\kern-.125emX}}

\definecolor{green}{rgb}{0.0, 0.5, 0.0} 
\definecolor{skyblue}{RGB}{100, 200, 255}
\definecolor{orange}{RGB}{255, 165, 50}
\begin{document}

\title{\textbf{CART-MPC}: \textbf{C}oordinating \textbf{A}ssistive Devices for \textbf{R}obot-Assisted \textbf{T}ransferring with Multi-Agent Model Predictive Control

% {\footnotesize \textsuperscript{*}Note: Sub-titles are not captured for https://ieeexplore.ieee.org  and
}

\author{
\IEEEauthorblockN{Ruolin Ye$^+$\IEEEauthorrefmark{1}, Shuaixing Chen$^+$\IEEEauthorrefmark{2}, Yunting Yan$^+$\IEEEauthorrefmark{1}, Joyce Yang\IEEEauthorrefmark{1}, Christina Ge\IEEEauthorrefmark{1}, Jose Barreiros\IEEEauthorrefmark{3}, Kate Tsui\IEEEauthorrefmark{3}, \\Tom Silver\IEEEauthorrefmark{1}, Tapomayukh Bhattacharjee\IEEEauthorrefmark{1}} 

\IEEEauthorblockA{\IEEEauthorrefmark{1}\textit{Cornell University}, Ithaca, NY, USA\\
\{ry273, yy2244, jby33, rg759, tss95, tapomayukh\}@cornell.edu} 
\IEEEauthorblockA{\IEEEauthorrefmark{2}\textit{Shanghai Jiao Tong University}, Shanghai, China\\
alkdischen@sjtu.edu.cn} 
\IEEEauthorblockA{\IEEEauthorrefmark{3}\textit{Toyota Research Institute}, Cambridge, MA, USA\\
\{jose.barreiros, kate.tsui\}@tri.global}
}

% \and
% \and
% \and
% \and
% \and

\twocolumn[{%
\renewcommand\twocolumn[1][]{#1}%

\maketitle
\thispagestyle{empty}
\pagestyle{empty}
\vspace{-2em}
\begin{center}
\begin{minipage}{\textwidth}
\centering
\includegraphics[width=0.95\textwidth]{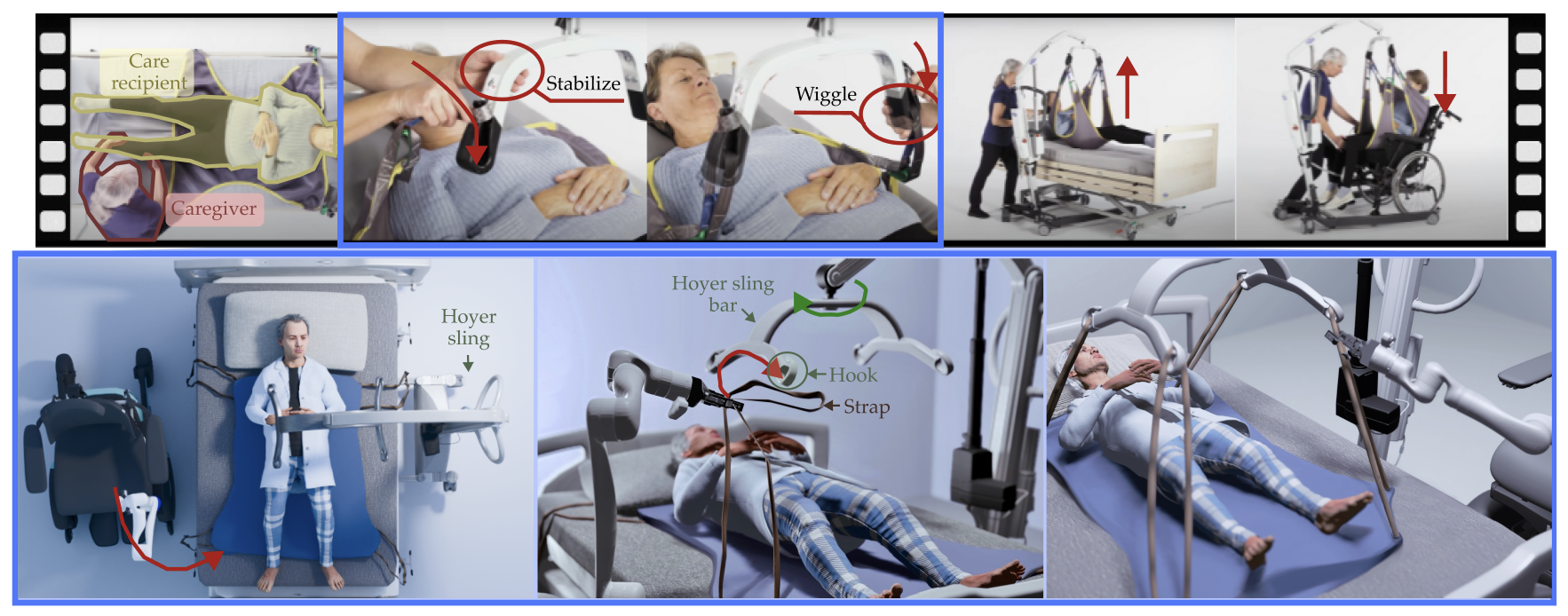}
\captionof{figure}{\footnotesize\textbf{Bed-to-wheelchair Transferring}: Human caregivers use assistive devices to perform transferring. For example, the caregiver in the figure (upper row) uses a Hoyer sling to move the care recipient from the hospital bed to the wheelchair. Inspired by this, we use an instrumented Hoyer sling along with a caregiving robot to perform the transferring task (lower row), where the robot performs fine manipulation, and the assistive devices take the heavy loads.}
\label{fig:teaser}
\end{minipage}
\end{center}

}]

\renewcommand{\thefootnote}{}% Remove numbering
\footnotetext{$^+$ denotes equal contribution. This work was partly funded by Toyota Research
Institute (TRI), National Science Foundation IIS \#2132846,and CAREER \#2238792. This article solely reflects the opinions and conclusions of its authors and not TRI or any other Toyota entity. We thank Xiaoman Yang for her help with rendering the teaser figure.}
\renewcommand{\thefootnote}{\arabic{footnote}}
\begin{abstract}
Bed-to-wheelchair transferring is a ubiquitous activity of daily living (ADL), but especially challenging for caregiving robots with limited payloads. We develop a novel \textcolor{black}{algorithm} that leverages the presence of other assistive devices: a Hoyer sling and a wheelchair for coarse manipulation of heavy loads, alongside a robot arm for fine-grained manipulation of deformable objects (Hoyer sling straps). We instrument the Hoyer sling and wheelchair with actuators and sensors so that they can become intelligent agents in the algorithm. We then focus on one subtask of the transferring ADL---tying Hoyer sling straps to the sling bar---that exemplifies the challenges of transfer: multi-agent planning, deformable object manipulation, and generalization to varying hook shapes, sling materials, and care recipient bodies. To address these challenges, we propose CART-MPC, a novel \textcolor{black}{algorithm} based on turn-taking multi-agent model predictive control that uses a learned neural dynamics model for a keypoint-based representation of the deformable Hoyer sling strap, and a novel cost function that leverages linking numbers from knot theory and neural amortization to accelerate inference. We validate it in both RCareWorld~\cite{ye2022rcare} simulation and real-world environments. In simulation, CART-MPC successfully generalizes across diverse hook designs, sling materials, and care recipient body shapes. In the real world, we show zero-shot sim-to-real generalization capabilities to tie deformable Hoyer sling straps on a sling bar towards transferring a manikin from a hospital bed to a wheelchair. See our website for supplementary materials: \href{https://emprise.cs.cornell.edu/cart-mpc/}{https://emprise.cs.cornell.edu/cart-mpc/}.
\end{abstract}

\section{Introduction}
Over 1 billion people worldwide require assistance with activities
of daily living (ADLs)~\cite{who2022}.
Recent work has demonstrated the promise of \emph{robot} assistance in ADLs such as feeding\textcolor{black}{~\cite{feeding1-flair, feeding2, feeding3, feeding4-repeat, feeding5-feelthebite, feeding6-askornot, feeling7-morpheus}}, bathing~\cite{bathing1,bathing2}, and dressing~\cite{dressing1,dressing2}.
However, transferring---helping a care recipient move between a bed and a wheelchair---has received relatively less attention in the physical human-robot interaction literature~\cite{riman,wang2014,kong2023}.
One reason is that transferring is very challenging given the limited torque capacity of collaborative robots.
For example, a Kinova robot arm with a 11.5 lb (5.2 kg) payload cannot independently move a 200 lb (90kg) care recipient.
Human caregivers, who also have limited physical strength, use assistive devices such as Hoyer slings\footnote{A Hoyer sling is a supportive device used with a mechanical lift (Hoyer lift) to safely transfer individuals with limited mobility; it consists of a wide, reinforced body support area with multiple attachment points (straps or loops) for securing to the lift.} and transfer boards to transfer care recipients.
These devices are very helpful, but they are also cumbersome to manipulate. 
In this work, we \emph{instrument} assistive devices with actuators and sensors so that they can be treated as robots themselves.
Together with a robot arm mounted on a wheelchair, this constitutes a set of instrumented assistive devices for transfer (Fig. \ref{fig:teaser}).

In this work, we focus on a subtask of transferring that exemplifies the key challenges of the overall task: \emph{fastening a Hoyer sling strap to the sling bar}.
In this subtask, the sling is initially positioned underneath a care recipient who is lying in bed.\footnote{Conversations with caregiving stakeholders suggest that the sling sheet can often remain underneath the care recipient while sleeping.}
The robot has grasped one of the sling straps and the sling bar is positioned overhead.
The goal of the subtask is to stably loop the strap into the hook.
Human caregivers typically use one hand to stabilize the bar to align it with the strap and then use another hand to fasten the strap on the hook (Fig.~\ref{fig:teaser}).
To accomplish this \emph{deformable object manipulation} with only one arm, we leverage the fact that the Hoyer sling bar itself is actuated \textcolor{black}{due to our instrumentation}.
However, coordinated planning of the robot arm \textcolor{black}{holding a strap} and sling bar has several major challenges: non-linear dynamics; infinite degrees of freedom; a 3D task space (\emph{c.f.} 2D rope manipulation~\cite{liu2023model}); and the instability of the rotating sling bar.
Furthermore, to be broadly useful, the algorithm must \emph{generalize} to different hook types, sling materials, and care recipient body shapes.

To address these challenges, we propose
\textbf{CART-MPC}: \textbf{C}oordinating \textbf{A}ssistive Devices for \textbf{R}obot-Assisted \textbf{T}ransferring with Multi-Agent \textbf{MPC}.
CART-MPC is a method for multi-agent model-predictive control (MPC) that uses learned dynamics in keypoint space with a novel strap-tying cost function.
Keypoints are sampled from segmentation using GroundedSAM2~\cite{ravi2024sam2segmentimages} and K-Medoids~\cite{rajivc2023segment}, and tracked using optical flow.
A keypoint-space neural dynamics model for the Hoyer sling strap is trained offline using data generated in the RCareWorld simulator~\cite{ye2022rcare}.
We leverage insights from knot theory~\cite{panagiotou2020knot} to create a cost function for strap tying.
The cost function is slow and challenging to compute from visual inputs, so we again use simulation data to train a fast neural amortization of it.
The dynamics model and cost functions are integrated in \textcolor{black}{a turn-taking} multi-agent MPC, coordinating the robot and Hoyer sling bar to fasten the Hoyer strap.

We evaluate CART-MPC in both simulation and the real world. We find that it generalizes to 5 different shapes of hooks, 3 different materials of the sling, and 5 different care recipients using the CareAvatars in RCareWorld~\cite{ye2022rcare} \textcolor{black}{validated using} 650 trials in total.
We further demonstrate that CART-MPC generalizes to the real world using a manikin. See the supplementary materials on our website: \href{https://emprise.cs.cornell.edu/cart-mpc/}{https://emprise.cs.cornell.edu/cart-mpc/}.

In summary, our contributions are as follows:
\begin{itemize}
    \item We propose CART-MPC, a novel method for \textcolor{black}{turn-taking} multi-agent model predictive control with learned keypoint-space dynamics.
    \item We instrument assistive devices and develop a new multi-agent platform, taking the first step towards autonomous bed-to-chair transferring.
    \item We propose a novel strap-tying cost function based on knot theory and train a neural network to amortize it.
    \item We show the generalizability of CART-MPC in RCareWorld, with 5 different shapes of hook, 3 different materials of the sling, and 5 different care recipients.
    \item We demonstrate the real-world applicability of CART-MPC in a manikin transferring scenario.
\end{itemize}
\section{Background}
Our work \textcolor{black}{on} \emph{robot-assisted transferring} (Sec.~\ref{Sec:Robot-Assisted-Transferring}) is inspired by \emph{human caregivers using assistive devices} (Sec. ~\ref{Sec:Assistive-Devices-for-Human-Assisted-Transferring}) for this task. Our algorithm is designed for \emph{multi-agent coordination in caregiving} (Sec. ~\ref{Sec:Multi-Agent-Coordination-in-Caregiving}). Specifically, we focus on tying the strap to the hooks, which is a task involving deformable object manipulation (Sec. ~\ref{Sec:Deformable-manipulation}).
\subsection{Robot-Assisted Transferring}\label{Sec:Robot-Assisted-Transferring}
Transferring is one of the less explored ADLs in the caregiving robotics literature.
Most relevant is early work on the Intelligent Sweet Home~\cite{zhang2014}, which proposed an intelligent hospital bed, Hoyer sling, and wheelchair designed for bed-to-wheelchair transfer.
However, to the best of our knowledge, this design was never fully realized and transferring was not demonstrated.
Furthermore, robot-assisted strap-tying was not considered, avoiding the deformable object manipulation challenges that we address in this work.
Previous work has also considered transferring with robot arms that have large payload capabilities.
For example, RI-MAN~\cite{riman}, ROBEAR~\cite{robear}, RIBA~\cite{riken2008riba}, RIBA-II~\cite{riken2011riba} use dual-arm humanoid robots with large payloads (up to 80 kg).
STRONGARM~\cite{wang2014} similarly features a large robot arm mounted to a wheelchair.
These approaches are limited by safety considerations, as well as by the large dimensions of the robots, and are not actively used in real caregiving settings.

Other work has considered custom-designed wheelchairs for transfer assistance, such as the HLPR chair~\cite{kaljanov2013}, which has a lift function.
These wheelchairs can be used for assisted sit-to-stand, but not for bed-to-wheelchair transfer.
There have also been efforts to develop hybrid wheelchair-bed devices that would reduce the need for transferring~\cite{mascaro1998docking,li2013design,singh2021design}.
Another line of work has considered the use of exoskeletons for transfer assistance.
Exoskeletons have been designed for human caregivers, reducing the physical effort required for human-assisted transferring~\cite{oconnor2021exoskeletons}, and also for care recipients themselves~\cite{exoskeleton2019carrecipient}.
Progress in this direction has been limited by safety concerns, fitting time, and exoskeleton speed~\cite{gorgey2018robotic}.
Overall, there remains a large gap between the transferring needs of care recipients and the robot-assisted solutions that are currently available.

\subsection{Assistive Devices for Human-Assisted Transferring}~\label{Sec:Assistive-Devices-for-Human-Assisted-Transferring}
Manual transferring remains common in human caregiving.
However, studies have found that 52\% of caregivers~\cite{kong2023} report occupational musculoskeletal pain due to the repetitive motion, awkward posturing, and heavy lifting involved in transferring.
Multiple assistive devices have been designed to mitigate these challenges.
Ceiling lifts~\cite{silvercross2023ceiling} raise care recipients and transfer them from point to point along tracks mounted on the ceiling.
While they are effective in transferring care recipients with severe mobility limitation, these lifts are expensive and require a permanent modification of the caregiving environments~\cite{pacificmobility2023ceiling}.

Hoyer slings feature passive wheels that allow mobility when pushed by a caregiver~\cite{wikipedia2023patient}. They are more affordable than other lifting solutions, making them accessible for care settings in both homes and caregiving facilities~\cite{caring2023hoyer}. However, Hoyer slings can still be cumbersome and time-consuming for caregivers to use.
In this work, we address this challenge by instrumenting a Hoyer sling with actuators and sensors, removing the need for human manipulation of the sling.

A wheelchair is a widely used assistive device designed to help transfer individuals from one location to another. These can be categorized into manual wheelchairs~\cite{manualwheelchair} and powered wheelchairs~\cite{smartwheelchair}. Powered wheelchairs are typically controlled by users through interfaces such as joysticks, but they often lack integration with other intelligent devices and the capability for autonomous movement. To address this limitation, we implemented a control interface that allows the wheelchair to receive commands from computers and equipped its mobile base with sensors to enable autonomous navigation through SLAM.

\subsection{Multi-Agent Coordination in Caregiving}~\label{Sec:Multi-Agent-Coordination-in-Caregiving}
Our instrumented Hoyer sling, wheelchair, and robot arm constitute a multi-agent system.
We propose a method for coordinating these agents to perform assisted transferring.
Multi-agent coordination has been considered in other caregiving paradigms such medicine management~\cite{Wagner2002} and health monitoring~\cite{intellcare}.
These systems typically feature a centralized autonomy allocation module that aggregates sensory input and sends commands to robots, virtual assistants, and/or human caregivers~\cite{Wagner2002, intellcare, Barber2022, marcon:hal-01518637, robocare}.
For instance, in HIMTAE~\cite{Barber2022} a centralized home-assistant unit allocates tasks to a dual arm manipulator along with other smaller mobile robots.
Klonovs et al. \cite{klonovs2015distributed} survey other distributed systems for monitoring older adults.
These systems typically involve passive sensing, rather than active robot decision-making.

Multi-agent coordination in general has several technical challenges~\cite{jager2001decentralized,amato2015planning,chen2021decentralized}.
A full review is outside the scope of this work.
For our purposes, the foremost challenge is decentralized planning to reach a common goal---fastening the Hoyer sling strap.
To address this challenge, we build on multi-agent MPC.
See~\cite{negenborn2009multi} for an early survey and~\cite{zhu2020trajectory,toumieh2022decentralized} for more recent examples.
To the best of our knowledge, multi-agent MPC has not been previously used in physical human-robot interaction.

\subsection{Deformable Object Manipulation in Caregiving}~\label{Sec:Deformable-manipulation}
Deformable object manipulation is a core challenge for robotics in general and for caregiving robotics in particular.
Nonlinear dynamics, infinite degrees of freedom, and partial occlusion are among the central difficulties~\cite{clothpoth, garmenttracking, unifolding, Wang2023One}.
Previous caregiving work has considered deformable object manipulation in ADLs such as feeding~\cite{ha2024repeat, feeding1-flair, feeding2, feeding3, ye2024morpheus} dressing (shirts~\cite{dressing1, dressing2}, pants~\cite{Yamazaki2014}, and shoes~\cite{Li2020}), and bathing (loofah and gel~\cite{bathing1}).
Instrumental ADLs such as laundry folding~\cite{Avigal2022SpeedFoldingLE} and bandaging~\cite{Li2023} also require deformable manipulation.

Planning for deformable object manipulation can be simulation-based \textcolor{black}{as well as} data-driven.
Simulation-based approaches use the material point method (MPM)~\cite{ha2024repeat}, extended particle-based dynamics~(XPBD)~\cite{ye2022rcare}, and incremental potential contact~(IPC) among other techniques to model the dynamics of deformable objects.
Data-driven approaches use learned dynamics models instead~\cite{yan2021learning,kim2022dsqnet,liu2023model}.
These approaches may use point clouds~\cite{Wang2023One}, RGB images~\cite{Finn2016}, or keypoints as inputs to the dynamics model~\cite{liu2023model}.
Building on both lines of work, our method models the Hoyer strap as a rope using XPBD in RCareWorld~\cite{ye2022rcare}, and leverages simulation-generated data to train a dynamics model in keypoint space.

\begin{figure}[ht!]
\centering
\includegraphics[width=1.0\linewidth]{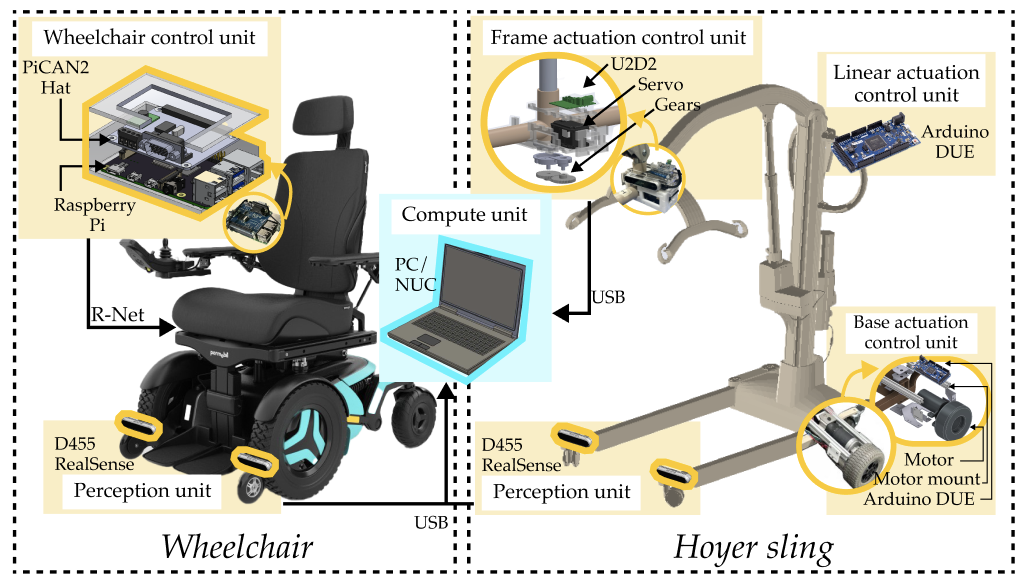}
\caption{
\footnotesize \textbf{Instrumented assistive devices:} We added sensors and actuators to a Hoyer sling and a commercial powered wheelchair with a robot arm. 
}
\vspace{-1em}
\label{fig:hardware}
\end{figure}
\section{Instrumented Assistive Devices for Transfer}
\label{sec:hardware}

In this section, we briefly describe our novel hardware setup to provide context for the technical contributions that are the focus of this work.
Further details can be found on our website.

% what are there
As illustrated in Figure~\ref{fig:hardware}, our instrumented devices consists of a wheelchair equipped with a Kinova Gen3 6-DoF robot arm and a Hoyer sling.
See Section~\ref{sec:experiment-setup} for hardware details.
Both the commercial wheelchair and Hoyer sling are actuated devices with limited autonomy. They offer control interfaces such as a joystick or handheld controller. The wheelchair offers 6 degrees of freedom, including features such as elevate (seat height), tilt (back + seat), recline (back only), and leg rest adjustments. In contrast, the Hoyer sling has a single degree of freedom, focused on lifting for care recipient transfer purposes.

Despite these capabilities, it is challenging to directly use these devices for our multi-agent algorithm for multiple reasons.
First, these devices lack control interfaces that have necessary communication protocols to receive commands from external systems such as a robot. While they have basic actuation for their primary functions, they do not have the level of actuation required to perform more complex tasks such as fastening the strap. Moreover, they lack sensing abilities to perceive the environment.

\textcolor{black}{
We instrument the assistive devices in three aspects: 
\textit{1) Actuation:}
We equip the Hoyer sling with 2 electric wheels, enabling it to perform mobile tasks. We also mount a servo on the sling bar, enabling it to rotate so that it can collaborate with the robot in the strap fastening task. 
\textit{2) Sensing:}
We mount two cameras on the mobile bases of each platform for navigation, enabling them to perform SLAM based on visual-inertial odometry.
We also mount two additional cameras on the sling bar---one on each side---to monitor the hooks.
\textit{3) Control interface:}
We implement a ROS-based control interface for the Hoyer sling and the wheelchair so that they can communicate with the robot. 
}
All together, these modifications enables the deployment of a multi-agent algorithm for intelligent coordination.

\section{The Hoyer Strap Fastening Task}

In this work, we focus on a challenging subtask of transferring: \textit{fastening a Hoyer sling strap to a hook on the sling bar} (see Figure \ref{fig:teaser}).
The task begins when the robot has the sling strap in the gripper and the bar is positioned overhead. 
The task is successful when the strap is tied to the hook on the bar.
\textcolor{black}{Our target population is} care recipients \textcolor{black}{with} severe mobility limitations, \textcolor{black}{thus} passive for the duration of the task.
The question of adapting to human movements or collaborating with active humans is left to future work.

Formally, our objective is to obtain a policy $\pi : \mathcal{O} \to \mathcal{A}$ that achieves a goal $g : \mathcal{O} \to \{\text{true}, \text{false}\}$:
\begin{itemize}
    \item An observation $o = (o_{\text{strap}}, o_{\text{bar}}) \in \mathcal{O}$ comprises RGB images $o_{\text{strap}}$ of the strap from two cameras---one on the sling bar and the other on the robot wrist---and the joint state $o_{\text{bar}} \in [0, 2\pi)$ of the sling bar.
    \item An action $a = (a_{\text{rob}}, a_{\text{bar}}) \in \mathcal{A}$ comprises a delta translation and rotation in the robot's end effector space $a_{\text{rob}} \in \mathbb{R}^6$ and rotation of the sling bar revolute joint $a_{\text{bar}} \in \mathbb{S}$ along a fixed axis.
    \item The goal $g(o)$ is true when the strap is stably fastened, meaning the strap does not fall within a fixed duration.
\end{itemize}
To achieve real-time decision-making under bandwidth constraints, we commit to a \emph{decentralized} policy: $$\pi(o) \triangleq (\pi_{\text{rob}}(o), \pi_{\text{bar}}(o))$$
where $\pi_{\text{rob}}(o) = a_{\text{rob}}$ and $\pi_{\text{bar}}(o) = a_{\text{bar}}$.
The combined policy is evaluated over multiple trials with variation in hook shapes, sling materials, and care recipient bodies.
An evaluation trial is considered successful if the goal $g$ is reached within a finite time horizon $T$.

\begin{figure*}[ht!]
\centering
\includegraphics[width=1.0\linewidth]{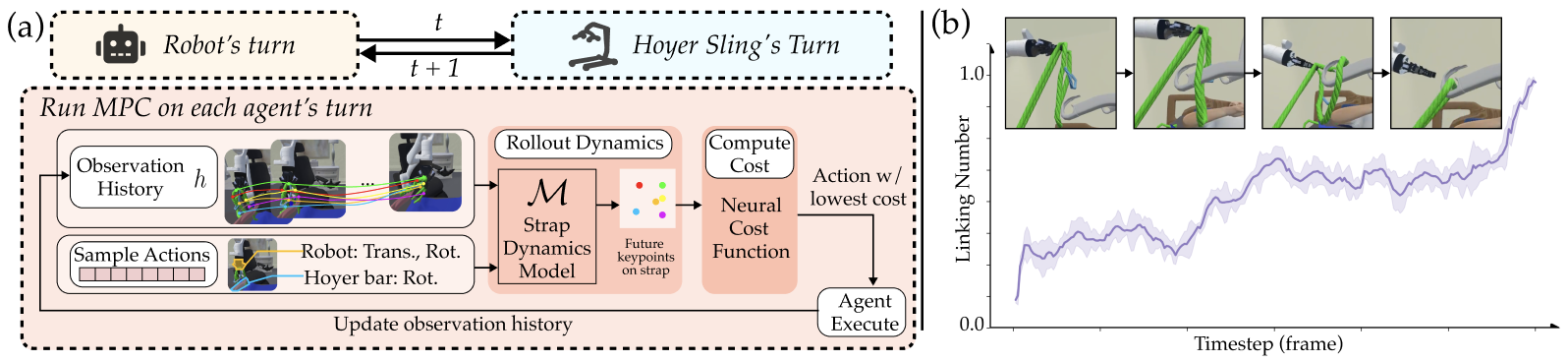}
\caption{\footnotesize (a) 
\textbf{CART-MPC}: We propose a turn-taking multi-agent algorithm to coordinate the robot and the Hoyer sling bar during strap fastening. 
The algorithm leverages a dynamics model for the strap and a neural cost function. See text for details.
(b) \textbf{Neural linking number}: We visualize the average neural linking number across five strap-tying trials. The neural linking number gradually increases from 0 to 1 as the robot ties the strap onto the hook.}
\vspace{-1em}
\label{fig:method}
\end{figure*}
\section{Method: Strap Fastening with CART-MPC}
We now present \textbf{CART-MPC}, our novel algorithm for the Hoyer strap fastening task.
See Figure~\ref{fig:method} for an overview of the algorithm.
CART-MPC is a decentralized policy $\pi$ with two parts: $\pi_{\text{rob}}$ for the robot and $\pi_{\text{bar}}$ for the Hoyer sling frame bar, where the \textcolor{black}{two agents take turns to perform the task.}
Both policies are implemented using multi-agent MPC (Sec.~\ref{subsec:mpc}).
Predictions in the MPC are governed by a learned dynamics model of the strap that operates in keypoint space (Sec.~\ref{subsec:dynamics}).
The MPC optimizes a weighted sum of costs, most notably including a novel knot-theory-based cost function (and a neural amortization of it) that represents the degree to which the strap is fastened on the hook  (Sec.~\ref{subsec:strap_tying_cost}).
We now describe each of these components in detail.

\subsection{Multi-agent Model Predictive Control}
\label{subsec:mpc}

CART-MPC uses turn-taking multi-agent MPC to coordinate the action of the robot and the Hoyer sling bar.
The robot and bar take turns: first, the robot policy $\pi_{\text{rob}}$ takes an action $a_{\text{rob}}$; then, the bar $\pi_{\text{bar}}$ takes an action $a_{\text{bar}}$.
Each policy selects its action by sampling candidates, predicting future states, and selecting the candidate action that minimizes expected future costs.
Both policies use a learned dynamics model (Sec.~\ref{subsec:dynamics}) for prediction and minimize the same cost function 
(Sec.~\ref{subsec:strap_tying_cost}).
For simplicity, each agent assumes that the other remains static during prediction.
We use $a^{\text{no}}_{\text{rob}}$ and $a^{\text{no}}_{\text{bar}}$ to denote no-change actions for the robot and bar respectively.
This process repeats until the goal $g$ is satisfied or the time horizon $T$ is reached.
See Alg.~\ref{alg:carema-mpc} for a summary.

\algnewcommand{\LineComment}[1]{\State \(\triangleright\) #1}

\begin{algorithm}
\caption{CART-MPC}
\label{alg:carema-mpc}
\begin{algorithmic}[1]

\Require Initial observation $o_0 = (o_{\text{strap}, 0}, o_{\text{bar}, 0})$, goal $g$, time horizon $T$, prediction model $\mathcal{M}$, cost function $C$.
\Ensure Actions $a_t = (a_{\text{rob}, t}, a_{\text{bar}, t})$ that achieve $g$.

\State \textbf{initialize} $t \gets 0$ and observation history $h = [~]$

\While{$g(o_t) = \text{false}$ and $t < T$}
    % \State \textbf{Robot's Turn:}  \textcolor{skyblack}{ $\rhd$ Sample and apply the robot action associated with the smallest cost for itself.}
    \LineComment \emph{Robot's Turn} ($\pi_{\text{rob,t}}$)
    \State Sample candidate actions $\mathcal{A}_\text{rob,t} = \{a_{\text{rob,t}}\}$
    \State Predict states $\hat{o}_{t+1} = \mathcal{M}(a_{\text{rob,t}}, a^{\text{no}}_{\text{bar,t}}, h_t)$ for $a_{\text{rob,t}} \in \mathcal{A}_\text{rob,t}$
    \State Execute $\arg \min_{a_{\text{rob,t}} \in \mathcal{A}_\text{rob,t}} C(a_{\text{rob,t}}, a^{\text{no}}_{\text{bar,t}}, h_t, \hat{o}_{t+1})$
    \State Update $t \gets t + 1$ and $h_t$
    \LineComment \emph{Hoyer Sling Bar's Turn} ($\pi_{\text{bar,t}}$)
    \State Sample candidate actions $\mathcal{A}_\text{bar,t} = \{a_{\text{bar,t}}\}$
    \State Predict states $\hat{o}_{t+1} = \mathcal{M}(a^{\text{no}}_{\text{rob,t}}, a_{\text{bar,t}}, h_t)$ for $a_{\text{bar,t}} \in \mathcal{A}_\text{bar,t}$
    \State Execute $\arg \min_{a_{\text{bar,t}} \in \mathcal{A}_\text{bar,t}} C(a^{\text{no}}_{\text{rob,t}}, a_{\text{bar,t}}, h_t, \hat{o}_{t+1})$
    \State Update $t \gets t + 1$ and $h_t$
\EndWhile

% \If{$g(o_t) = \text{true}$}
%     \State Return "Task Successful"
% \Else
%     \State Return "Task Failed"
% \EndIf

\end{algorithmic}
\end{algorithm}

\subsection{Dynamics: Learning a Model in Keypoint Space}
\label{subsec:dynamics}

We now describe the prediction model $\mathcal{M}$, which takes in the observation history $h_t = (o_0, o_1 \dots, o_t)$, a robot action, and a bar action, and predicts a future observation:
$$\hat{o}_{t+1} = \mathcal{M}(a_{\text{rob},t}, a_{\text{bar},t}, h_t).$$

We assume that the sling bar update is fully determined by the bar action, that is, $\hat{o}_{\text{bar}, t+1} = f(\hat{o}_{\text{bar}, t}, a_{\text{bar}, t})$ is known.
The more challenging problem is predicting the strap $\hat{o}_{\text{strap}, t+1}$.

\subsubsection{Strap State Space}

To address this problem, we first need to determine a representation for the state of the strap that can be used for effective planning.
Inspired by previous work in deformable object manipulation~\cite{liu2023model}, we choose a \emph{keypoint-based representation} of the strap state.
This representation contains sufficient information for planning and can be efficiently computed from visual inputs.

We now detail the strap keypoint representation.
For simplicity, the following discussion describes the process of computing keypoints for one camera; in practice, we compute keypoints for both cameras and join them together.
Given an initial RGB image of the strap \textcolor{black}{$o_{\text{strap}, t}$}, we start by performing instance segmentation of the strap using GroundedSAM2~\cite{ravi2024sam2segmentimages}.
We then use K-Medoids~\cite{rajivc2023segment} to sample $n$ initial keypoints $k^i_{0} \in \mathbb{R}^2$, for $i=1, \dots, n$, within the segmentation mask.\footnote{The keypoints are in 2D (in the pixel space). The cameras have depth functionality, but they are too close to the hook to have meaningful depth information when the strap is interacting with the hook.} 

Upon receiving future camera images $o_{\text{strap}, t+1}$, we update the keypoints using the current keypoints $k_t = \{k^i_t\}_{i=1}^n$ and the current image $o_{\text{strap}, t}$.
The two images are compared using the Lucas-Kanade method to create an optical flow map $F_{t} : \mathbb{R}^2 \to\mathbb{R}^2$.
Each keypoint is updated following the optical flow: $k^i_{t+1} = k^i_{t} + F_{t}(k^i_{t})$.
After updating, we check to see whether each keypoint is within the new segmentation mask computed by GroundedSAM2 on $o_{\text{strap}, t+1}$.
If a keypoint is outside the mask, we resample a nearby point within the mask.
Specifically, we find the closest mask point to the keypoint using Euclidean distance, draw a small circle around that point, and sample from the intersection of the circle and mask.
\subsubsection{Learning a Dynamics Model}
We next learn a dynamics model in keypoint space to complete the prediction model $\mathcal{M}$.
To train the dynamics model, we generate trajectory data using the RCareWorld simulator~\cite{ye2022rcare}, sampling trajectories of keypoints $k_t$ and bar observations $o_{\text{bar}, t}$ using a random exploration policy.
We use the known simulator state to compute exact keypoints, rather than tracking them visually.
From these data, we learn a neural dynamics model $\mathbf{D}$:
$$k_{t+1} = \mathbf{D}(k_0, o_{\text{bar}, 0}, a_0, \dots, k_t, o_{\text{bar}, t}, a_t)$$
In practice, we use only the $t = 20$ most recent observations in $h_t$ and parameterize $\mathbf{D}$ as an MLP with 7 layers and train it by optimizing MSE loss using Adam~\cite{kingma2014adam}.
See our website for hyperparameters and other details.

\subsection{Cost Function: Strap Tying with Linking Numbers}
\label{subsec:strap_tying_cost}

When planning with the dynamics model, we optimize a weighted sum of cost functions.
In this section, we describe a novel \emph{strap tying} cost that measures the extent to which the Hoyer strap is fastened around the sling bar hook.

\subsubsection{Preliminaries}

Our strap tying cost function takes inspiration from knot theory.
The \emph{linking number} $\operatorname{link}(\gamma_1, \gamma_2)$ of two closed curves $\gamma_1$ and $\gamma_2$ is an integer that counts the number of times that one curve crosses the other~\cite{panagiotou2020knot}.
We are interested in the case where one curve---the Hoyer strap---is closed and the other---the sling bar hook---is open.
In this case, $\operatorname{link}(\gamma_1, \gamma_2)$ is no longer an integer, but the intuition is the same: the linking number represents the degree to which one curve winds around the other~\cite{panagiotou2020knot}.

\subsubsection{Computing Linking Number in Simulation}
We use linking number to measure the extent to which the Hoyer strap is fastened around the sling bar hook.
In simulation, we have access to ground-truth pose and shape information and can associate discrete 3D points to parameterize a curve on each of the objects.
Let $\mathbf{\gamma}^i_1 \in \mathbb{R}^3$ denote the $i^{th}$  point on the Hoyer strap and $\mathbf{\gamma}^j_2 \in \mathbb{R}^3$ be the $j^{th}$ point on the sling bar hook.
\textcolor{black}{
Let $\operatorname{link_\Sigma}(\gamma_1, \gamma_2)$ be a discrete version of the linking number $\operatorname{link}(\gamma_1, \gamma_2)$, we then compute it using
}
$$
\frac{1}{4 \pi} \sum_{i=1}^{M_1-1} \sum_{j=1}^{M_2-1} \frac{\left( \mathbf{\gamma}_1^i - \mathbf{\gamma}_2^j \right) \cdot \left( (\mathbf{\gamma}_1^{i+1} - \mathbf{\gamma}_1^i) \times (\mathbf{\gamma}_2^{j+1} - \mathbf{\gamma}_2^j) \right)}{|\mathbf{\gamma}_1^i - \mathbf{\gamma}_2^j|^3}
$$
where $M_1$ and $M_2$ are the numbers of \textcolor{black}{key points} sampled on the Hoyer strap and sling bar hook respectively.

\begin{figure*}[ht!]
\centering
\includegraphics[width=1.0\linewidth]{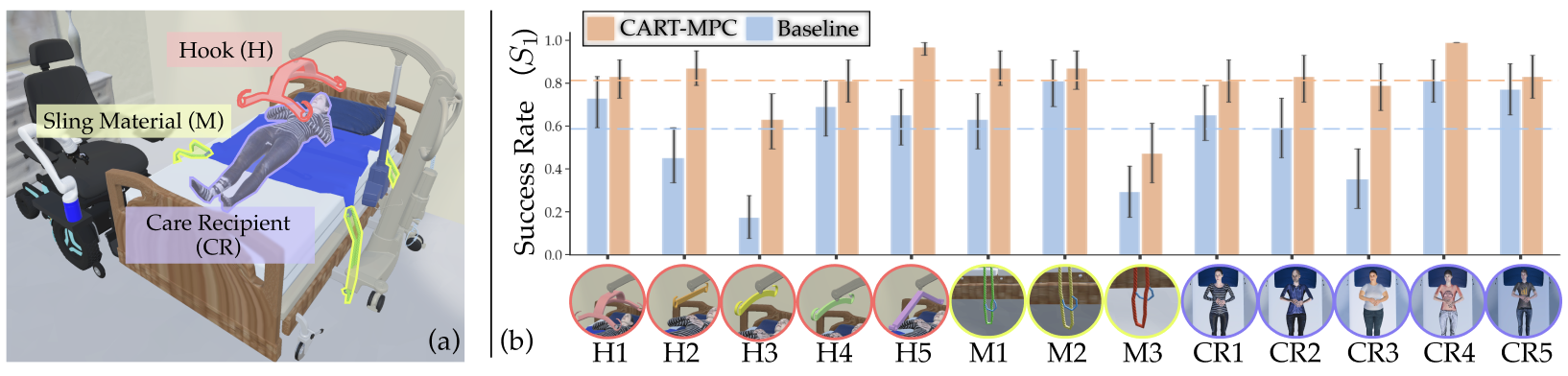}
\caption{\footnotesize
\textbf{Setup and results for evaluation of CART-MPC \textcolor{black}{in RCareWorld}}: We evaluate our method using the setup on the left with various Hook shapes (H), Care Recipients (CR) \textcolor{black}{with various body shapes}, and Sling Materials (M) \textcolor{black}{with three levels of compliance}. We show the results $S_1$ in (b) comparing CART-MPC with a single-agent MPC with a passive sling bar. The results suggest that CART-MPC using multi-agent performs better than a single-agent baseline method. See Table~\ref{tab:main-results} for full results.}
\vspace{-2em}
\label{fig:eval-setup-result}
\end{figure*}

\subsubsection{Neural Amortization}
The discrete linking number $\operatorname{link}_\Sigma(\gamma_1, \gamma_2)$ requires ground-truth pose information and is slow to run during planning $(M_1, M_2 > 1000)$. We detail the speed-accuracy balance on the website.
We therefore use data offline generated in simulation to train a neural network that maps observations to linking number.
Using a random exploration policy, we generate $m$ simulation states and evaluate $\operatorname{link}_\Sigma$ in each.
We then train an LSTM (parameters detailed on the website) to map full observation histories to linking numbers. 
This results in a neural approximation $\operatorname{link}_\theta(h_t) \in \mathbb{R}$ of the linking number that can be efficiently evaluated given the information available during planning. Figure~\ref{fig:method} (b) illustrates the linking number predicted by the neural network as the robot ties the strap onto the hook. 

\subsubsection{Final Cost}
We use the neural amortized linking number to define a strap typing cost function $$c_{\text{link}}(h_t) = 1 - \frac{\operatorname{link}_\theta(h_t) - \beta_0}{\beta_1-\beta_0}$$ where $\beta_0$ and $\beta_1$ are the minimum and maximum values respectively of $\operatorname{link}_\theta$ observed in the training data.
The cost $c_{\text{link}}(h_t)$ is therefore between 0 and 1 and achieves lower values when the Hoyer strap is more tightly fastened around the sling bar hook.

This cost function, together with the learned dynamics function and multi-agent MPC algorithm, complete CART-MPC as a method for the strap-tying task.

\begin{figure*}[ht!]
\centering
\includegraphics[width=1.0\linewidth]{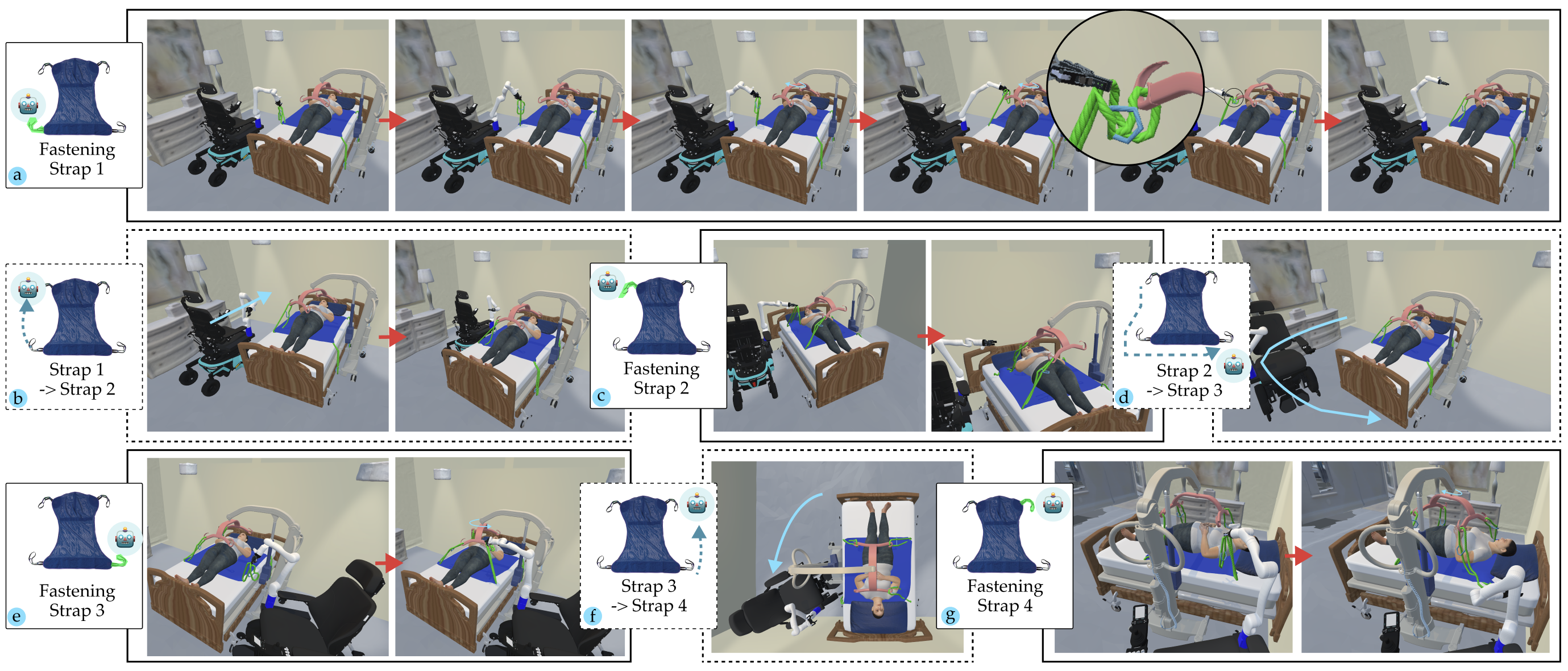}
\caption{\footnotesize
\textbf{CART-MPC executed sequence in RCareWorld}: We demonstrate one executed sequence with H1, M1, and CR3. In this trial, the robot and the sling bar collaboratively fasten the 4 straps to the sling bar hooks.}
\label{fig:sim-exec}
\end{figure*}

\begin{table*}[t]
\centering
\setlength{\tabcolsep}{6pt}  
\caption{\small Evaluating CART-MPC: We compare our method with the baseline methods w/o multi-agent collaboration and demonstrate that our method generalizes to different hook shapes, sling materials, and care recipient body shapes. \textcolor{black}{All trials with 4 straps successfully tied on lead to successful lifting.}}
\label{tab:main-results}
\begin{tabular}{c|c|*{5}{>{\centering\arraybackslash}p{0.5cm}}|*{3}{>{\centering\arraybackslash}p{1.0cm}}|*{5}{>{\centering\arraybackslash}p{0.5cm}}}
\hline
\multicolumn{1}{c}{\multirow{2}{*}{}} & & \multicolumn{5}{c|}{Various H, fixed M \& CR} & \multicolumn{3}{c|}{Various M, fixed H \& CR} & \multicolumn{5}{c}{Various CR, fixed H \& M} \\
\cline{3-15}
\multicolumn{1}{c}{} & & $H1$ & $H2$ & $H3$ & $H4$ & $H5$ & $M1$ & $M2$ & $M3$ & $CR1$ & $CR2$ & $CR3$ & $CR4$ & $CR5$ \\
\hline\hline
\multirow{3}{*}{$S_1$} & Baseline-uncontrolled & 0.74& 0.46	& 0.18	&0.7	&0.66&	0.64	&0.82	&0.3	&0.66&	0.6&	0.36	&0.82&	0.78 \\
& Baseline-fixed & 0.62&	0.4	&0.18	&0.76	&0.58	&0.66	&0.62	&0.22	&0.68	&0.46	&0.48	&0.84	&0.58 \\
& CART-MPC &\textbf{0.84}	&\textbf{0.88}	&\textbf{0.64}	&\textbf{0.82}	&\textbf{0.98}	&\textbf{0.88}	&\textbf{0.88}	&\textbf{0.48}	&\textbf{0.82}	&\textbf{0.84}	&\textbf{0.8}	&\textbf{1}&	\textbf{0.84} \\
\hline
\multirow{3}{*}{$S_4$} & Baseline-uncontrolled & 0.82	&0.62	&0.34	&0.66	&0.72	&\textbf{0.8}	&0.64	&0.42	&0.8	&0.68	&0.46	&0.8	&0.66 \\
& Baseline-fixed & 0.78	& 0.48	&0.42	&0.64	&0.68	&0.78	&0.8	&0.34	&0.78	&0.6	&0.4&\textbf{0.8}	&0.62 \\
& CART-MPC & \textbf{0.84}	&\textbf{0.72}	&\textbf{0.7}	&\textbf{0.84}	&\textbf{0.94}	&\textbf{0.8}	&\textbf{0.8}	&\textbf{0.44}	&\textbf{0.86}	&\textbf{0.8}	&\textbf{0.82}	&\textbf{0.88}	&\textbf{0.8} \\
\hline
\multirow{3}{*}{$S_{lift}$} & Baseline-uncontrolled & 0.82	&0.62	&0.34	&0.66	&0.72	&\textbf{0.8}	&0.64	&0.42	&0.8	&0.68	&0.46	&0.8	&0.66 \\
& Baseline-fixed & 0.78	& 0.48	&0.42	&0.64	&0.68	&0.78	&0.8	&0.34	&0.78	&0.6	&0.4&\textbf{0.8}	&0.62 \\
& CART-MPC & \textbf{0.84}	&\textbf{0.72}	&\textbf{0.7}	&\textbf{0.84}	&\textbf{0.94}	&\textbf{0.8}	&\textbf{0.8}	&\textbf{0.44}	&\textbf{0.86}	&\textbf{0.8}	&\textbf{0.82}	&\textbf{0.88}	&\textbf{0.8} \\
\hline
\end{tabular}
\vspace{-3mm}
\end{table*}

\section{Experiments}
We perform experiments to evaluate the effectiveness of CART-MPC.

\noindent We first describe the experimental setup and summarize the results (Table~\ref{tab:main-results}). \textcolor{black}{We then perform significance tests to validate one hypothesis and discuss other two findings.} Check our website for videos of CART-MPC running in simulation and the real world.

\subsection{Experimental Setup}
\label{sec:experiment-setup}

\subsubsection{Hardware}
As described in Sec.~\ref{sec:hardware}, our hardware setup includes a wheelchair, a robot arm, and a Hoyer sling.
The wheelchair is a Rovi x3\cite{rovix3}, the robot arm is a Kinova Gen3 6-DoF\cite{kinovagen3} with a Robotiq 85 gripper\cite{robotiq85}, and the Hoyer sling is an Invacare Reliant 450\cite{invacare450}.
We use a DYNAMIXEL MX-64T servo\cite{dynamixel} to actuate the rotation of the sling bar.
The navigation cameras are RealSense D515\cite{d515}; the sling bar cameras are RealSense D415\cite{d415}; and the robot wrist camera is a RealSense D435\cite{d435}.
We use Arduino\cite{arduino} Due as the microcontroller for the Hoyer sling and Raspberry Pi\cite{raspberrypi} for the wheelchair.

\subsubsection{Compute}
Our system's core operations are powered by an AMD Ryzen 9 5900X CPU with a base clock speed of 3.7 GHz. For tasks requiring accelerated computation, such as simulations and neural network inference, we rely on an NVIDIA GeForce RTX 3090 GPU with 24 GB of GDDR6X memory. For enhanced performance, the system is supported by 32 GB of DDR4 RAM. Additionally, we use two laptops for the wheelchair and the Hoyer sling to handle local control, e.g., for navigation.
All machines run Ubuntu 20.04.
\subsubsection{Controllers}
We implement multi-agent MPC using STORM~\cite{bhardwaj2022storm}, which is a sampling-based MPC implemented on GPU. The MPC runs at approximately 30 Hz.
We detail the cost terms derived from STORM and the corresponding weights on our website.

\subsubsection{Simulation} 
We use RCareWorld~\cite{ye2022rcare} as the simulation platform. We set the time step to 0.02s to make the soft body simulation smooth. We simulate the straps as 1D particle chains (ropes), and the sling body as a 2D particle mesh (a cloth sheet) using the Obi physics backend in RCareWorld. We attach the straps to the sling body using pin constraints. We model the hooks and the sling bar as a time-varying signed distance function~(SDF) for faster and more robust collision detection.
In simulation, we consider 5 types of hooks \textcolor{black}{(C-shaped and variations)},
3 types of sling materials (compliant, medium, stiff), and 5 care recipients with various body shapes from RCareWorld~\cite{ye2022rcare}, visualized in Fig.~\ref{fig:eval-setup-result}.

\subsection{Evaluating with Additional Transfer Components}
While our focus in this paper is on the Hoyer strap tying subtask, we aim to evaluate our method in the context of the broader transferring task. 
We consider two evaluation schemes: \emph{single-strap tying} and \emph{full-sling tying}.
For single-strap tying, the robot starts from a predefined position with the strap already in its gripper.
For full-sling tying, four positions for the wheelchair are predefined, with the assumption that the wheelchair's control system can accurately follow a planner-generated trajectory. \textcolor{black}{This assumption and the evaluation process are detailed in Sec.~\ref{sec:full-demo}}
In simulation, the robot's gripper is directly attached to a specific point on the strap, while in real-world trials, we ensure the straps are consistently placed in the same position before each test so that grasping success is guaranteed. 

In all cases, we count a fastening attempt as a \emph{success} if the strap stays securely at the hook's bottom. 
In single-strap tying, the robot attempts strap tying only once per trial; in full-sling tying, the robot can attempt up to three times to secure each strap.
We use $S_1$ and $S_4$ to denote the success rates for single-strap tying and full-sling tying, respectively. We also report $S_{lift}$, the success rate of lifting the care recipient after securing 4 straps; here, the care recipient must remain stable in the sling for at least 5 seconds post-lift.

We compare CART-MPC with two baselines in RCareWorld.
Both baselines are single-agent, as opposed to CART-MPC, which is multi-agent.
The first baseline treats the sling bar as a passive but rotatable object; the second baseline treats the sling bar as fixed, without any rotation.
We evaluate all methods in simulation across 5 hook shapes, 3 sling materials, and 5 care recipient body shapes. Each setting is repeated for 50 trials for a total of $50 \cdot (5 + 3 + 5) = 650$ trials per method. Notably, the dynamics model is only trained with sling material $M_1$ and hook shape $H_1$.

We show the simulation environment setup in Fig.~\ref{fig:eval-setup-result} with a visualization of the results for $S_1$ and 
the full results in Table~\ref{tab:main-results}.

\begin{figure*}[ht!]
\centering
\includegraphics[width=1.0\linewidth]{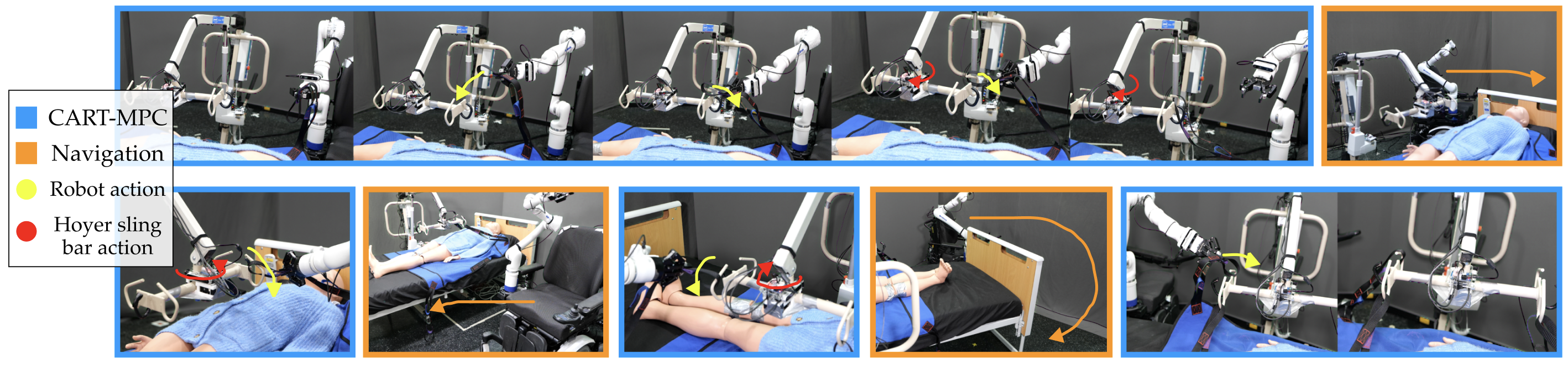}
\caption{\footnotesize
\textbf{CART-MPC execute sequence in the real world}: We demonstrate one execute sequence with a manikin. The robot and the Hoyer sling bar collaboratively fasten four straps. See our website for the full video.}
\vspace{-1em}
\label{fig:real-exec}
\end{figure*}

\subsection{Evaluating the Effectiveness of CART-MPC}
\vspace{-0.48em}
\subsubsection{Hypothesis: By leveraging multi-agent coordination, CART-MPC outperforms single-agent MPC baselines with uncontrolled or fixed sling bars}
Single-agent MPC with an uncontrolled sling bar achieves an average $S_1 = 0.59$, $S_4 = 0.64$, and $S_{lift} = 0.64$  compared to $S_1 = 0.54$, $S_4 = 0.62$, and $S_{lift} = 0.62$ with a fixed sling bar, while CART-MPC achieves $S_1 = 0.82$, $S_4 = 0.78$, and $S_{lift} = 0.78$. Statistical analysis using a z-test shows significant differences in success rates ($p < 0.0001$). 
This supports our hypothesis that multi-agent coordination is crucial for strap tying.

Qualitatively, we observed that the \textcolor{black}{sling bar movement} provides added flexibility for hook-rope interactions by offering a \textcolor{black}{passive/active} rotational degree of freedom, leading to a slightly higher success rate. For instance, when the strap rests on the tip of the hook, and the robot pulls on it, a movable hook reduces the likelihood of the strap slipping out, making the uncontrolled sling bar more effective than a fixed one.
However, active control of the sling bar proves more efficient and effective, supporting our hypothesis.
    
\subsubsection{CART-MPC generalizes across hook shapes, sling materials, and care recipient body shapes}
We evaluate the performance of CART-MPC \textcolor{black}{trained on $M_1$ and $H_1$}, and compare it with other hook shapes and sling materials to assess its generalizability. 
\textcolor{black}{The results in Table~\ref{tab:main-results} suggest that all setups have similar performances with the training set, except for $M_3$, showing the generalizability of CART-MPC.}

Qualitatively, we see that the linking-number-inspired cost function, which leverages keypoint positions on the hook, allows it to be effectively applied to various hook shapes as long as they share the same topology, \textcolor{black}{i.e., a curve segment with an opening.}
We also note that different care recipient body shapes introduce varying levels of slack in the sling sheet and strap. For example, a wider body may press against the sling sheet’s edges, reducing slack and making strap tying more challenging.
The sling bar’s rotation addresses this issue by adjusting the hook’s position closer to the strap, enhancing robustness across diverse body shapes.
We do observe a performance drop on $M_3$, likely due to the dynamics model being trained exclusively on $M_1$. While this enabled reasonable generalization to $M_2$, additional training with diverse materials could further improve performance across a broader range of sling materials such as $M_3$.

\subsubsection{CART-MPC generalizes to the real world}\label{sec:full-demo}
We demonstrate the real-world applicability of CART-MPC in a manikin transferring scenario.
We use the dynamics model obtained from simulation and transfer it directly to the real world, but tune the MPC parameters separately for the real-world scenario. \textcolor{black}{See our website for full video demonstrations.}

\textcolor{black}{\textbf{Single-strap tying:}}
We evaluated real-world performance with 50 single-strap tying trials using a Hoyer sling and a manikin, achieving a 62\% success rate.

This sim-to-real success is made possible by several design choices.
First, by using GroundedSAM2 for keypoint detection, we do not need to collect real-world training data to handle real image inputs.
Second, the wiggling motion of the sling bar found by CART-MPC increases the error tolerance of the robot policy.
Third, the MPC prediction horizon is sufficiently short, and the learned dynamics model is sufficiently general that the distribution shift between sim and real is not a bottleneck. We also tune the material in the simulation to be aligned with the real world as much as possible to reduce the sim-to-real gap.
Finally, the neural amortization of our strap-tying cost function allows us to efficiently guide MPC using only the keypoint representation that can be obtained in the real world.

\textcolor{black}{\textbf{Full demonstration:}} We demonstrate a full strap-tying process (Fig.~\ref{fig:real-exec}) leveraging the navigation functionality of the wheelchair. The wheelchair starts from a predefined position and navigates itself next to the first strap using RTAB-Map~\cite{Labbe2019} to pick up the strap. It moves the strap to a start position slightly above the hook using an RRT planner. It runs CART-MPC to tie the strap on the hook. Then, the process for the rest of the straps is repeated. 

\section{Discussion}
There remain several open challenges for both the strap-tying subtask and assisted transferring as a whole.
\textcolor{black}{Despite the impressive generalizability on various hook shapes and care recipient body shapes, the neural dynamics model has limited ability to generalize to out-of-distribution sling materials.}
Multi-agent coordination mitigates this to some degree---the sling bar's wiggle motion can increase the error tolerance for the robot policy---but the challenge remains. We envision training on a more diverse set of sling materials will potentially improve generalizability. 
Another area for future work is improving the fidelity of the simulator we use to generate training data. 
However, simulating interactions between thin objects and soft-rigid systems, such as straps and hooks, remains a challenging problem.
Also, the linking number-based cost function can sometimes produce falsely high values, resulting in tying failures—particularly when the strap is very close to the hook but not yet secured. We believe this issue arises because the input is limited to a 2D pixel space. Incorporating 3D information, such as depth data, may help address this limitation.

In assisted transferring, there are other important subtasks to consider: positioning the sling underneath the care recipient if it is not already there; safely executing bed-to-wheelchair transfer after the straps are attached; and preparing moving wheelchair for wheelchair-to-bed transfer, among others. We aim to look into these interesting while challenging problems using the multi-agent system in the future.

\bibliographystyle{ieeetr}
\balance
\bibliography{references}

\begin{thebibliography}{10}

\bibitem{ye2022rcare}
R.~Ye, W.~Xu, H.~Fu, R.~K. Jenamani, V.~Nguyen, C.~Lu, K.~Dimitropoulou, and T.~Bhattacharjee, ``{RCareWorld}: A human-centric simulation world for caregiving robots,'' in {\em 2022 IEEE/RSJ International Conference on Intelligent Robots and Systems (IROS)}, pp.~33--40, IEEE, 2022.

\bibitem{who2022}
W.~H. Organization, {\em Global report on health equity for persons with disabilities}.
\newblock World Health Organization, 2022.

\bibitem{feeding1-flair}
R.~K. Jenamani, P.~Sundaresan, M.~Sakr, T.~Bhattacharjee, and D.~Sadigh, ``Flair: Feeding via long-horizon acquisition of realistic dishes,'' {\em arXiv preprint arXiv:2407.07561}, 2024.

\bibitem{feeding2}
E.~K. Gordon, R.~K. Jenamani, A.~Nanavati, Z.~Liu, D.~Stabile, X.~Dai, T.~Bhattacharjee, T.~Schrenk, J.~Ko, H.~Bolotski, {\em et~al.}, ``An adaptable, safe, and portable robot-assisted feeding system,'' in {\em Companion of the 2024 ACM/IEEE International Conference on Human-Robot Interaction}, pp.~74--76, 2024.

\bibitem{feeding3}
J.~Ondras, A.~Anwar, T.~Wu, F.~Bu, M.~Jung, J.~J. Ortiz, and T.~Bhattacharjee, ``Human-robot commensality: Bite timing prediction for robot-assisted feeding in groups,'' in {\em 6th Annual Conference on Robot Learning}, 2022.

\bibitem{feeding4-repeat}
H.~Nayoung, Y.~Ruolin, L.~Ziang, S.~Shubhangi, and B.~Tapomayukh, ``Repeat: A real2sim2real approach for pre-acquisition of soft food items in robot-assisted feeding,'' 2024.

\bibitem{feeding5-feelthebite}
R.~K. Jenamani, D.~Stabile, Z.~Liu, A.~Anwar, K.~Dimitropoulou, and T.~Bhattacharjee, ``Feel the bite: Robot-assisted inside-mouth bite transfer using robust mouth perception and physical interaction-aware control,'' in {\em ACM/IEEE International Conference on Human Robot Interaction (HRI)}, 2024.

\bibitem{feeding6-askornot}
R.~Banerjee, R.~K. Jenamani, S.~Vasudev, A.~Nanavati, S.~Dean, and T.~Bhattacharjee, ``To ask or not to ask: Human-in-the-loop contextual bandits with applications in robot-assisted feeding,'' 2024.

\bibitem{feeling7-morpheus}
Y.~Ruolin, H.~Yifei, B.~Yuhan, K.~Luke, and B.~Tapomayukh, ``Morpheus: a multimodal one-armed robot-assisted peeling system with human users in-the-loop,'' 2024.

\bibitem{bathing1}
R.~Madan, S.~Valdez, D.~Kim, S.~Fang, L.~Zhong, D.~T. Virtue, and T.~Bhattacharjee, ``Rabbit: A robot-assisted bed bathing system with multimodal perception and integrated compliance,'' in {\em Proceedings of the 2024 ACM/IEEE International Conference on Human-Robot Interaction}, pp.~472--481, 2024.

\bibitem{bathing2}
C.-H. King, T.~L. Chen, A.~Jain, and C.~C. Kemp, ``Towards an assistive robot that autonomously performs bed baths for patient hygiene,'' in {\em 2010 IEEE/RSJ International Conference on Intelligent Robots and Systems}, pp.~319--324, IEEE, 2010.

\bibitem{dressing1}
A.~Jevti{\'c}, A.~F. Valle, G.~Aleny{\`a}, G.~Chance, P.~Caleb-Solly, S.~Dogramadzi, and C.~Torras, ``Personalized robot assistant for support in dressing,'' {\em IEEE transactions on cognitive and developmental systems}, vol.~11, no.~3, pp.~363--374, 2018.

\bibitem{dressing2}
J.~Zhu, M.~Gienger, G.~Franzese, and J.~Kober, ``Do you need a hand?--a bimanual robotic dressing assistance scheme,'' {\em IEEE Transactions on Robotics}, vol.~40, pp.~1906--1919, 2024.

\bibitem{riman}
{RIKEN-TRI Collaboration Center for Human-Interactive Robot Research}, ``Ri-man,'' 2006.

\bibitem{wang2014}
H.~Wang, C.-Y. Tsai, H.~Jeannis, C.~S. Chung, A.~Kelleher, G.~G. Grindle, and R.~A. Cooper, ``Stability analysis of electrical powered wheelchair-mounted robotic-assisted transfer device,'' {\em Journal of Rehabilitation Research and Development}, vol.~51, no.~5, pp.~761--774, 2014.

\bibitem{kong2023}
Y.-K. Kong, K.-H. Choi, S.-S. Park, J.-W. Shim, and H.-H. Shim, ``Evaluation of the efficacy of a lift-assist device regarding caregiver posture and muscle load for transferring tasks,'' {\em International Journal of Environmental Research and Public Health}, vol.~20, no.~2, p.~1174, 2023.

\bibitem{liu2023model}
Z.~Liu, G.~Zhou, J.~He, T.~Marcucci, L.~Fei-Fei, J.~Wu, and Y.~Li, ``Model-based control with sparse neural dynamics,'' in {\em Thirty-seventh Conference on Neural Information Processing Systems (NeurIPS)}, (Vancouver, Canada), December 2023.

\bibitem{ravi2024sam2segmentimages}
N.~Ravi, V.~Gabeur, Y.-T. Hu, R.~Hu, C.~Ryali, T.~Ma, H.~Khedr, R.~Rädle, C.~Rolland, L.~Gustafson, E.~Mintun, J.~Pan, K.~V. Alwala, N.~Carion, C.-Y. Wu, R.~Girshick, P.~Dollár, and C.~Feichtenhofer, ``Sam 2: Segment anything in images and videos,'' 2024.

\bibitem{rajivc2023segment}
F.~Raji{\v{c}}, L.~Ke, Y.-W. Tai, C.-K. Tang, M.~Danelljan, and F.~Yu, ``Segment anything meets point tracking,'' {\em arXiv preprint arXiv:2307.01197}, 2023.

\bibitem{panagiotou2020knot}
E.~Panagiotou and L.~H. Kauffman, ``Knot polynomials of open and closed curves,'' {\em Proceedings of the Royal Society A}, vol.~476, no.~2240, p.~20200124, 2020.

\bibitem{zhang2014}
C.-Q. Zhang, Y.-F. Liu, and D.-P. Liang, ``Realization of ageing-friendly smart home system with computational intelligence,'' {\em Journal of Computational Information Systems}, vol.~10, no.~8, pp.~3583--3590, 2014.

\bibitem{robear}
{RIKEN-TRI Collaboration Center for Human-Interactive Robot Research}, ``Robear,'' 2015.

\bibitem{riken2008riba}
{RIKEN-TRI Collaboration Center for Human-Interactive Robot Research}, ``Riba: Robot for interactive body assistance,'' 2008.

\bibitem{riken2011riba}
RIKEN-TRI, ``Riken-tri collaboration center for human-interactive robot research,'' 2011.
\newblock Accessed: 2024-09-26.

\bibitem{kaljanov2013}
R.~Bostelman, J.~Albus, and J.~Johnson, ``Hlpr chair: a novel patient transfer device,'' in {\em Proceedings of the 8th Workshop on Performance Metrics for Intelligent Systems}, PerMIS '08, (New York, NY, USA), p.~302–305, Association for Computing Machinery, 2008.

\bibitem{mascaro1998docking}
S.~Mascaro and H.~H. Asada, ``Docking control of holonomic omnidirectional vehicles with applications to a hybrid wheelchair/bed system,'' in {\em Proceedings. 1998 IEEE International Conference on Robotics and Automation (Cat. No. 98CH36146)}, vol.~1, pp.~399--405, IEEE, 1998.

\bibitem{li2013design}
F.~Li, C.~Zhang, H.~Liu, L.~Gao, J.~Ye, and D.~Xin, ``Design of a new multifunctional wheelchair-bed,'' in {\em World Congress on Medical Physics and Biomedical Engineering May 26-31, 2012, Beijing, China}, pp.~1342--1345, Springer, 2013.

\bibitem{singh2021design}
S.~Singh, S.~N. Panda, R.~K. Kaushal, N.~Kumar, and J.~L. Raheja, ``Design and development of iot enabled hybrid wheelchair cum bed,'' in {\em 2021 International Conference on Emerging Smart Computing and Informatics (ESCI)}, pp.~711--715, IEEE, 2021.

\bibitem{oconnor2021exoskeletons}
S.~O'Connor, ``Exoskeletons in nursing and healthcare: A bionic future,'' {\em Clinical Nursing Research}, vol.~30, pp.~1123--1126, Nov 2021.
\newblock Epub 2021 Aug 8.

\bibitem{exoskeleton2019carrecipient}
J.~Khan, K.~M.~R. Songlap, A.~Mizan, M.~Sahar, and S.~Ahmed, ``Assistive exoskeleton for paralyzed people,'' pp.~474--479, 01 2019.

\bibitem{gorgey2018robotic}
A.~S. Gorgey, ``Robotic exoskeletons: The current pros and cons,'' {\em World Journal of Orthopedics}, vol.~9, pp.~112--119, Sep 18 2018.

\bibitem{silvercross2023ceiling}
S.~Cross, ``Ceiling lifts,'' 2023.
\newblock Accessed: September 26, 2024.

\bibitem{pacificmobility2023ceiling}
P.~Mobility, ``Ceiling lift vs. floor lift: Pros and cons,'' 2023.
\newblock Accessed: September 26, 2024.

\bibitem{wikipedia2023patient}
W.~contributors, ``Patient lift,'' 2023.
\newblock Accessed: September 26, 2024.

\bibitem{caring2023hoyer}
Caring.com, ``Guide to choosing the best hoyer lift,'' 2023.
\newblock Accessed: September 26, 2024.

\bibitem{manualwheelchair}
B.~Dudgeon, J.~Deitz, and M.~Dimpfel, {\em Wheelchair Selection}, pp.~495--579.
\newblock 01 2014.

\bibitem{smartwheelchair}
R.~Simpson, ``Smart wheelchairs: A literature review,'' {\em Journal of rehabilitation research and development}, vol.~42, pp.~423--36, 07 2005.

\bibitem{Wagner2002}
T.~A. Wagner, J.~Phelps, M.~Scheutz, and T.~Bauer, ``Achieving global coherence in multi-agent caregiver systems: Centralized versus distributed response coordination in i.l.s.a.,'' in {\em Proceedings of the AAAI Conference on Artificial Intelligence}, 2002.

\bibitem{intellcare}
P.~Valente, S.~Hossain, B.~Grønbæk, K.~Hallenborg, and L.~P. Reis, ``A multi-agent framework for coordination of intelligent assistive technologies,'' in {\em 5th Iberian Conference on Information Systems and Technologies}, pp.~1--6, 2010.

\bibitem{Barber2022}
R.~Barber, F.~J. Ortiz, S.~Garrido, F.~M. Calatrava-Nicolás, A.~Mora, A.~Prados, J.~A. Vera-Repullo, J.~Roca-González, I.~Méndez, and O.~M. Mozos, ``A multirobot system in an assisted home environment to support the elderly in their daily lives,'' {\em Sensors}, vol.~22, no.~20, p.~7983, 2022.

\bibitem{marcon:hal-01518637}
E.~Marcon, S.~Chaabane, Y.~Sallez, T.~Bonte, and D.~Trentesaux, ``{A multi-agent system based on reactive decision rules for solving the caregiver routing problem in home health care},'' {\em {Simulation Modelling Practice and Theory}}, vol.~74, pp.~134--151, May 2017.
\newblock IF=1.954.

\bibitem{robocare}
S.~Bahadori, A.~Cesta, G.~Grisetti, L.~Iocchi, G.~R. Leone, D.~Nardi, A.~Oddi, F.~Pecora, and R.~Rasconi, ``Robocare: an integrated robotic system for the domestic care of the elderly,'' 01 2003.

\bibitem{klonovs2015distributed}
J.~Klonovs, M.~A. Haque, V.~Krueger, K.~Nasrollahi, K.~Andersen-Ranberg, T.~B. Moeslund, and E.~G. Spaich, {\em Distributed computing and monitoring technologies for older patients}.
\newblock Springer, 2015.

\bibitem{jager2001decentralized}
M.~Jager and B.~Nebel, ``Decentralized collision avoidance, deadlock detection, and deadlock resolution for multiple mobile robots,'' in {\em Proceedings 2001 IEEE/RSJ International Conference on Intelligent Robots and Systems. Expanding the Societal Role of Robotics in the the Next Millennium (Cat. No. 01CH37180)}, vol.~3, pp.~1213--1219, IEEE, 2001.

\bibitem{amato2015planning}
C.~Amato, G.~Konidaris, G.~Cruz, C.~A. Maynor, J.~P. How, and L.~P. Kaelbling, ``Planning for decentralized control of multiple robots under uncertainty,'' in {\em 2015 IEEE international conference on robotics and automation (ICRA)}, pp.~1241--1248, IEEE, 2015.

\bibitem{chen2021decentralized}
Y.~Chen, U.~Rosolia, and A.~D. Ames, ``Decentralized task and path planning for multi-robot systems,'' {\em IEEE Robotics and Automation Letters}, vol.~6, no.~3, pp.~4337--4344, 2021.

\bibitem{negenborn2009multi}
R.~R. Negenborn, B.~De~Schutter, and J.~Hellendoorn, ``Multi-agent model predictive control: A survey,'' {\em arXiv preprint arXiv:0908.1076}, 2009.

\bibitem{zhu2020trajectory}
E.~L. Zhu, Y.~R. St{\"u}rz, U.~Rosolia, and F.~Borrelli, ``Trajectory optimization for nonlinear multi-agent systems using decentralized learning model predictive control,'' in {\em 2020 59th IEEE Conference on Decision and Control (CDC)}, pp.~6198--6203, IEEE, 2020.

\bibitem{toumieh2022decentralized}
C.~Toumieh and A.~Lambert, ``Decentralized multi-agent planning using model predictive control and time-aware safe corridors,'' {\em IEEE Robotics and Automation Letters}, vol.~7, no.~4, pp.~11110--11117, 2022.

\bibitem{clothpoth}
W.~Xu, W.~Du, H.~Xue, Y.~Li, R.~Ye, Y.-F. Wang, and C.~Lu, ``Clothpose: A real-world benchmark for visual analysis of garment pose via an indirect recording solution,'' in {\em Proceedings of the IEEE/CVF International Conference on Computer Vision (ICCV)}, pp.~58--68, October 2023.

\bibitem{garmenttracking}
H.~Xue, W.~Xu, J.~Zhang, T.~Tang, Y.~Li, W.~Du, R.~Ye, and C.~Lu, ``Garmenttracking: Category-level garment pose tracking,'' in {\em Proceedings of the IEEE/CVF Conference on Computer Vision and Pattern Recognition (CVPR)}, pp.~21233--21242, June 2023.

\bibitem{unifolding}
H.~Xue, Y.~Li, W.~Xu, H.~Li, D.~Zheng, and C.~Lu, ``Unifolding: Towards sample-efficient, scalable, and generalizable robotic garment folding,'' in {\em {CoRL} Conference on Robot Learning}, 2023.

\bibitem{Wang2023One}
Y.~Wang, Z.~Sun, Z.~Erickson, and D.~Held, ``One policy to dress them all: Learning to dress people with diverse poses and garments,'' in {\em Robotics: Science and Systems (RSS)}, 2023.

\bibitem{ha2024repeat}
N.~Ha, R.~Ye, Z.~Liu, S.~Sinha, and T.~Bhattacharjee, ``Repeat: A real2sim2real approach for pre-acquisition of soft food items in robot-assisted feeding,'' in {\em IEEE International Conference on Intelligent Robots and Systems}, 2024.

\bibitem{ye2024morpheus}
R.~Ye, Y.~Hu, Y.~Bian, L.~Kulm, and T.~Bhattacharjee, ``{MORPHeus: a Multimodal One-armed Robot-assisted Peeling system with Human Users in-the-loop},'' in {\em International Conference on Robotics and Automation}, 2024.

\bibitem{Yamazaki2014}
K.~Yamazaki, R.~Oya, K.~Nagahama, K.~Okada, and M.~Inaba, ``Bottom dressing by a life-sized humanoid robot provided failure detection and recovery functions,'' in {\em Proceedings of the 2014 IEEE/SICE International Symposium on System Integration}, (Tokyo, Japan), December 13--15 2014.

\bibitem{Li2020}
Y.~Li, A.~Xiao, Q.~Feng, T.~Zou, and C.~Tian, ``Design of service robot for wearing and taking off footwear,'' in {\em E3S Web of Conferences}, vol.~189, p.~03024, 2020.

\bibitem{Avigal2022SpeedFoldingLE}
Y.~Avigal, L.~Berscheid, T.~Asfour, T.~Kroger, and K.~Goldberg, ``Speedfolding: Learning efficient bimanual folding of garments,'' {\em 2022 IEEE/RSJ International Conference on Intelligent Robots and Systems (IROS)}, pp.~1--8, 2022.

\bibitem{Li2023}
J.~Li, W.~Sun, X.~Gu, J.~Guo, J.~Ota, Z.~Huang, and Y.~Zhang, ``A method for a compliant robot arm to perform a bandaging task on a swaying arm: A proposed approach,'' {\em IEEE Robotics and Automation Magazine}, vol.~30, pp.~50--61, 2023.

\bibitem{yan2021learning}
W.~Yan, A.~Vangipuram, P.~Abbeel, and L.~Pinto, ``Learning predictive representations for deformable objects using contrastive estimation,'' in {\em Conference on Robot Learning}, pp.~564--574, PMLR, 2021.

\bibitem{kim2022dsqnet}
S.~Kim, T.~Ahn, Y.~Lee, J.~Kim, M.~Y. Wang, and F.~C. Park, ``Dsqnet: a deformable model-based supervised learning algorithm for grasping unknown occluded objects,'' {\em IEEE Transactions on Automation Science and Engineering}, vol.~20, no.~3, pp.~1721--1734, 2022.

\bibitem{Finn2016}
C.~Finn, I.~Goodfellow, and S.~Levine, ``Unsupervised learning for physical interaction through video prediction,'' in {\em Advances in Neural Information Processing Systems}, vol.~29, 2016.

\bibitem{kingma2014adam}
D.~P. Kingma, ``Adam: A method for stochastic optimization,'' {\em arXiv preprint arXiv:1412.6980}, 2014.

\bibitem{rovix3}
{Rovi Mobility}, ``Rovi x3 power wheelchair,'' 2024.
\newblock Accessed: 2024-01-05.

\bibitem{kinovagen3}
{Kinova Inc.}, {\em Gen3 Ultra lightweight robot - Technical Documentation}.
\newblock Kinova Robotics, 2019.
\newblock Accessed: 2024-01-05.

\bibitem{robotiq85}
{Robotiq Inc.}, ``2f-85 adaptive robot gripper,'' 2023.

\bibitem{invacare450}
{Invacare Corporation}, ``Invacare reliant 450 electric patient lift,'' 2024.
\newblock Accessed: 2024-01-05.

\bibitem{dynamixel}
{ROBOTIS}, ``Dynamixel mx-64t,'' 2023.

\bibitem{d515}
{Intel Corporation}, ``Intel realsense lidar camera l515,'' 2023.

\bibitem{d415}
{Intel Corporation}, ``Intel realsense depth camera d415,'' 2023.

\bibitem{d435}
{Intel Corporation}, ``Intel realsense depth camera d435,'' 2023.

\bibitem{arduino}
{Arduino}, ``Arduino due,'' 2023.

\bibitem{raspberrypi}
{Raspberry Pi Foundation}, ``Raspberry pi,'' 2023.

\bibitem{bhardwaj2022storm}
M.~Bhardwaj, B.~Sundaralingam, A.~Mousavian, N.~D. Ratliff, D.~Fox, F.~Ramos, and B.~Boots, ``{STORM}: An integrated framework for fast joint-space model-predictive control for reactive manipulation,'' in {\em Conference on Robot Learning}, pp.~750--759, PMLR, 2022.

\bibitem{Labbe2019}
M.~Labbé and F.~Michaud, ``{RTAB-Map} as an open-source lidar and visual simultaneous localization and mapping library for large-scale and long-term online operation,'' {\em Journal of Field Robotics}, vol.~36, no.~2, pp.~416--446, 2019.

\end{thebibliography}

\end{document}